\documentclass[journal]{IEEEtran}
\usepackage{cite}
\usepackage{color}
\usepackage{times}
\usepackage{soul}
\usepackage{url}
\usepackage[hidelinks]{hyperref}
\usepackage[utf8]{inputenc}
\usepackage[small]{caption}
\usepackage{graphicx}
\usepackage{amsmath}
\usepackage{amsthm}
\usepackage{booktabs,bigstrut}
\usepackage{algorithm}
\usepackage{algorithmic}
\graphicspath{{Figure/}}
\usepackage{multirow}
\usepackage{multicol}
\usepackage[american]{babel}
\usepackage{microtype}
\usepackage{subfigure}

\begin{document}
	
	\title{Efficient and Model-Based Infrared and Visible Image Fusion via Algorithm Unrolling}
	\author{Zixiang Zhao,
			Shuang Xu,
			Jiangshe Zhang,
			Chengyang Liang,
			Chunxia Zhang,
			Junmin Liu,~\IEEEmembership{Member,~IEEE}
		\thanks{The research is supported by the National Key Research and Development Program of China under grant 2020AAA0105601, the National Natural Science Foundation of China under grant 61976174 and 61877049. \textit{(Corresponding author: Jiangshe Zhang).}}
		\thanks{The authors are with the School of Mathematics and Statistics, Xi’an Jiaotong University, Xi’an, Shaanxi, 710049, P.R.China (E-mail: \{zixiangzhao, shuangxu, liangcy1996\}@stu.xjtu.edu.cn, \{jszhang, cxzhang, junminliu\}@mail.xjtu.edu.cn).}
		\thanks{Copyright © 2021 IEEE. Personal use of this material is permitted. However, permission to use this material for any other purposes must be obtained from the IEEE by sending an email to pubs-permissions@ieee.org.}
	}
	
	\markboth{IEEE Transactions on Circuits and Systems for Video Technology}%
	{Z.X. Zhao \MakeLowercase{\textit{et al.}}: Efficient and Interpretable Infrared and Visible Image Fusion Via Algorithm Unrolling}
	
	\maketitle
	
	\begin{abstract}
		Infrared and visible image fusion (IVIF) expects to obtain images that retain thermal radiation information from infrared images and texture details from visible images.
		In this paper, a model-based convolutional neural network (CNN) model, referred to as Algorithm Unrolling Image Fusion (AUIF), is proposed to overcome the shortcomings of traditional CNN-based IVIF models. 
		The proposed AUIF model starts with the iterative formulas of two traditional optimization models, which are established to accomplish two-scale decomposition, i.e., separating low-frequency base information and high-frequency detail information from source images.
		Then the algorithm unrolling is implemented where each iteration is mapped to a CNN layer and each optimization model is transformed into a trainable neural network.
		Compared with the general network architectures, the proposed framework combines the model-based prior information and is designed more reasonably.
		After the unrolling operation, our model contains two decomposers (encoders) and an additional reconstructor (decoder). In the training phase, this network is trained to reconstruct the input image. While in the test phase, the base (or detail) decomposed feature maps of infrared/visible images are merged respectively by an extra fusion layer, and then the decoder outputs the fusion image.
		Qualitative and quantitative comparisons demonstrate the superiority of our model, which can robustly generate fusion images containing highlight targets and legible details, exceeding the state-of-the-art methods. Furthermore, our network has fewer weights and faster speed.
	\end{abstract}
	
	
	\IEEEpeerreviewmaketitle

	\section{Introduction}\label{sec:1}
	\IEEEPARstart{I}{mage} fusion, as an image enhancement technology, is a hot issue in the image processing research community. By merging the images obtained by different sensors in the same scene, we expect to obtain images that highlight the advantages of each source image and are robust to perturbations at the same time~\cite{ma2019infrared,8653405,DBLP:conf/aaai/JingLDWDSW20,DBLP:journals/tci/XuJWLSZZ20,DBLP:journals/tgrs/XuALZZL20}.
	Image fusion can effectively improve the utilization of image information, eliminate conflicts and redundancies among multiple sensors, while form a clear and complete description of targets to facilitate recognition and tracking in subsequence \cite{meher2019a,7855783,zhao2020bayesian}. Infrared and visible image fusion, abbreviated as IVIF, is a typical topic in image fusion~\cite{9310288,9335976,9349250,zhaoijcai2020}. By incorporating prior knowledge to the images during the preprocessing stage, IVIF is effective to make full use of information in images and widely used in fire control~\cite{lahoud2018ar}, autonomous driving~\cite{li2020ivfusenet,4318244} and face recognition~\cite{ma2016infrared}, etc.
	\begin{figure}[t]
		\centering
		\includegraphics[width=\linewidth]{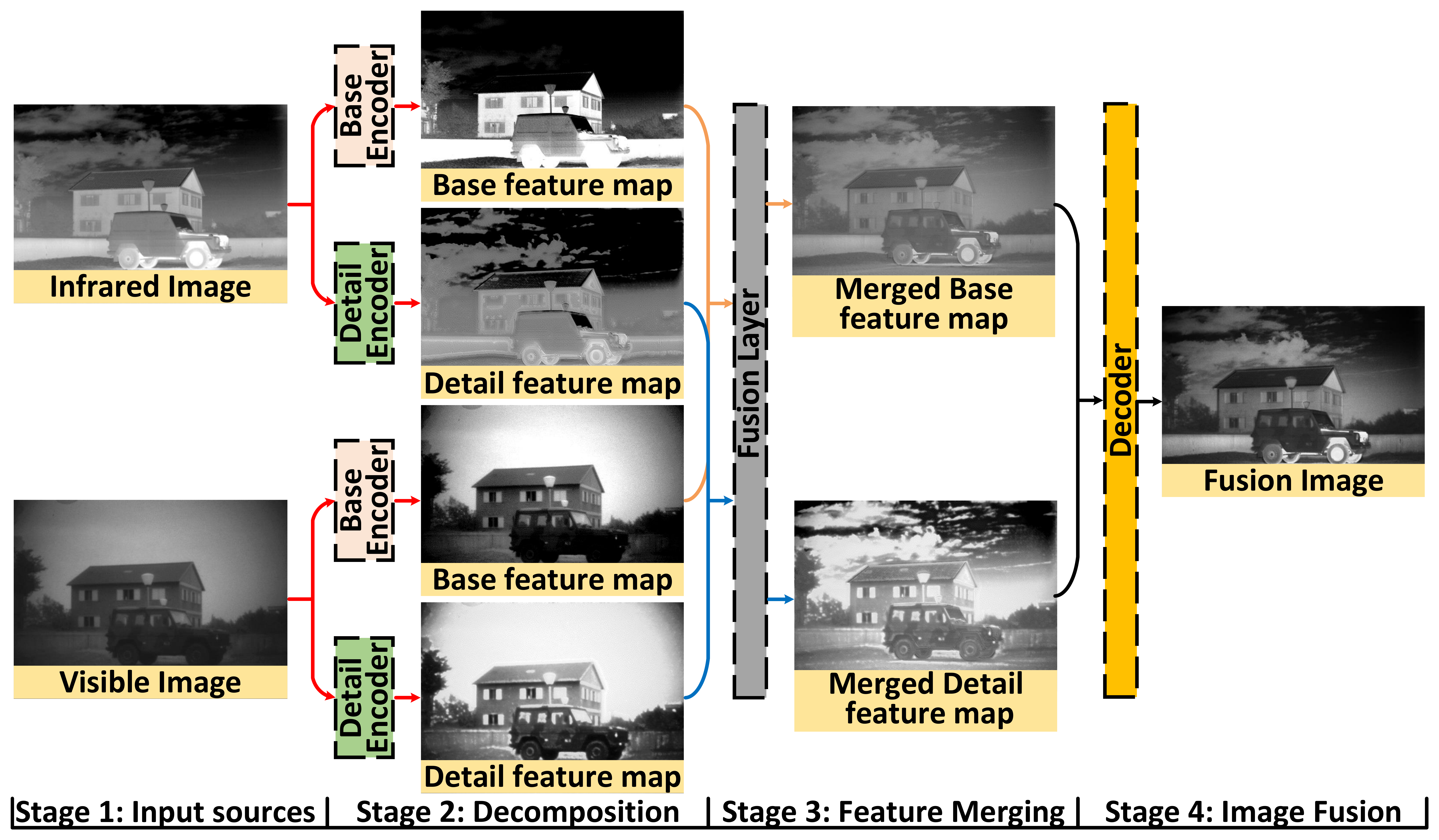}
		\caption{Workflow of the proposed deep network fusion model.}
		\label{fig:Workflow}
	\end{figure}
	
	Commonly, the infrared image is used to characterize the heat of objects, which is strongly robust to illumination changes and artifacts. Targets in the infrared image are usually highlighted and easy to distinguish from the environment.
	However, the texture and gradient information is seriously lost, and the spatial resolution is low. Hence, it is difficult to make satisfactory descriptions of the target details.
	In contrast, the visible image is good at keeping the information of gradient and pixel intensity, and displaying the brightness of objects. The content and objects can be described with enriched details and high resolution.
	However, as it is easily affected by illumination changes and light reflection, objects are difficult to be extracted from the background.
	IVIF aims at generating fusion images with both detailed texture information and highlight radiation information so as to form a clear, complete and accurate description of the targets, which is significant for the tracking and identification image tasks \cite{ma2016infrared}.
	
	Recent IVIF algorithms can be divided into classic methods and deep learning (DL)-based methods.
	Multi-scale decomposition (MSD) is one of the most promising techniques among classic methods. It usually separates an image into multiple-level images based on some criteria and uses a specific merging strategy to fuse the separated images at different levels. Finally, the fusion image can be obtained by adding the decomposed images of each level together \cite{li2011performance}.
	Among the decomposition methods, filters \cite{toet1989image} and transformers (e.g., wavelet\cite{liu2014region} and curvelet\cite{choi2005fusion}) are the most frequently employed. The difficulty of MSD is how to design reasonable decomposition algorithms and fusion strategies.
	
	With the development of DL in the field of computer vision, deep neural networks (DNNs) have been widely used in the IVIF task due to their strong feature extraction capability.
	DL-based methods can be divided into three categories: pre-trained model class, generative adversarial net (GAN) class, and auto-encoder (AE) class.
	The first class is the combination of MSD and DL \cite{li2018infrared,lahoud2019fast}. After MSD, the base images are weighted averaged. Then, the detail images with high-frequency information are fused by a pre-trained neural network (such as VGG-19 \cite{simonyan2014very}).
	The second class is the GAN-based method.
	In FusionGAN \cite{ma2019fusiongan}, the image fusion task is described as an adversarial game. The generator generates an image with the advantages of two source images. The discriminator adds detail information to the fusion image by forcing the generator to output the image similar to the source visible image. Recently, DPAL \cite{ma2020infrared} improves the quality of detailed information in fusion images by means of detail loss and target edge-enhancement loss.
	The third class is to train an AE network, in which the encoder and decoder are responsible for feature extraction and image reconstruction. For example, Densefuse \cite{li2018densefuse} trains an AE network with dense blocks \cite{huang2017densely}. In the test phase, the feature maps of source images extracted by the encoder are fused by weighted-average, and the fused image is obtained by the decoder. In general, DL is more effective than classic methods for strong feature extraction ability.
	
	It is noteworthy that the existing DL-based methods often stack convolutional neural layers empirically. As we know, DNN is a kind of black-box nonlinear transformation that could be learned through data-driven methods. Hence, the meaning of network structure and transformations are not clear. Conversely, in this paper, by the principle of algorithm unrolling, we extend an optimization-based two-scale decomposition algorithm into a {model-driven} DNN, named as Algorithm Unrolling Image Fusion (AUIF). The workflow of the AUIF model is presented in Fig.~\ref{fig:Workflow}. Our contributions are as follows:
	
	(1) We propose an efficient and {model-based} IVIF network. Initially, two traditional optimization models are established to accomplish two-scale decomposition, separating low-frequency base information and high-frequency detail information from source images. By the principle of algorithm unrolling, the fixed and manually-designed filters (or say the matrix multiplication) in the iterative formula are replaced with learnable neural network units. In this fashion, the update steps are mapped to our proposed AUIF network, which makes its structure easier to understand. Additionally, {the parameters in the proposed AE model} can be trained end-to-end by the back-propagation algorithm, {thus it is obvious that our network alleviates the influence of hyperparameters in the original optimization problem.} In short, our model is more effective than traditional MSD-based fusion methods, and meanwhile more rational than DL-based fusion methods.
	
	(2) We give a more comprehensive test of algorithm performance. The current methods \cite{ma2016infrared,li2018densefuse,zhang2017infrared,li2018infrared} use only a part of images in the TNO dataset for testing, where the scene is limited to the nightlight illumination. To make a convincing evaluation, we employ 132 test images from TNO, NIR, and FLIR datasets with diverse scenes. The complicated illumination conditions and various objects make the test scenario more comprehensive. Compared with eight state-of-the-art (SOTA) algorithms, the qualitative and quantitative results on the test datasets imply that our method is the best performer and it can stably generate fusion images with sharpening edges and abundant details in all three datasets. In addition, the reproducibility and computational efficiency of our model are also tested and the superiority of them are proved.
	
	The rest of this paper consists of the following sections. We briefly review the related work in section~\ref{sec:2}. Then in section~\ref{sec:3}, the formulation and implementation of our model are introduced. The results and analyses of numerous experiments are reported in section~\ref{sec:4}. Finally, we give conclusions in section~\ref{sec:5}.
	
	\section{Related Work}\label{sec:2}
	\subsection{Multi-scale Decomposition}
	Many IVIF algorithms have been proposed, and MSD is one of the most promising techniques. Its basic idea is to decompose the original picture into a group of images, each of which contains unique characters. Instead of fusing the original pictures, MSD fuses the decomposed images, and the fused pictures are recovered by inverse MSD. The popular decomposition methods include pyramid transform~\cite{BULANON200912}, discrete cosine transform \cite{JIN20181}, nonsubsampled contourlet transform \cite{XIANG201553} and bilateral filter \cite{HU2012196}.
	
	Two classes of DL-based methods are closely related to MSD. The first one incorporates pre-trained DNNs in an image fusion pipeline. But it is found that this technique is less effective. The other one is the AE networks based method. Actually, the decomposed images can be regarded as feature maps, while MSD and inverse MSD correspond to encoder and decoder, respectively. Therefore, in the era of DL, MSD is progressively replaced by AE networks. The representative models are dense block-based AE~\cite{li2018densefuse} and U-net's variant~\cite{DBLP:journals/corr/abs-1905-11447}. Compared with manually designed MSD methods, the data-driven AE networks are more flexible but the underlying structures are difficult to expound.
	\subsection{Algorithm Unrolling}
	Recently, an emerging technique called algorithm unrolling provides an encouraging pipeline to design {model-based} DNNs. One of the seminal works is the fast sparse coding proposed by Gregor and LeCun \cite{DBLP:conf/icml/GregorL10}. The traditional sparse coding problem is solved by iterative algorithms. And the idea of algorithm unrolling is to extend the iterative algorithm's computational graph into a DNN, in which the pre-defined hyperparameters and unknown coefficients can be trained end-to-end.
	
	Similar works have appeared in other low-level vision fields. To solve the blind image deblurring task, Li et al.~\cite{li2020efficient} unroll the iterative algorithm for the gradient total-variation regularization model and achieve significant
	practical performance. For the single image super-resolution task, Zhang et al.~\cite{zhang2020deep} unfold the maximum a posteriori probability (MAP) estimate model via a half-quadratic splitting algorithm and enhance the flexibility of the proposed network to different scale factors or noise levels.	Overall, the algorithm unrolling based DL is very competitive but with high rationality and less number of parameters~\cite{DBLP:journals/corr/abs-1912-10557}.
	
	\section{Method}\label{sec:3}
	This section formulates a new image two-scale decomposition model and introduces how this model is unrolled to a neural network. The details of AUIF architecture are also mentioned.
	\subsection{Motivation}
	Among the current IVIF algorithms, it is difficult for the classic methods to separate the low-frequency base information and high-frequency detail information by means of simple filters, manually-designed optimization models and transformers.
	{As for most of the DL-based methods, they are data-driven and black-box models which are designed empirically.} Thus, their working mechanisms remain unclear and inefficient.
	
	Accordingly, we propose an optimization-based image decomposition model to decompose an input image into a base image and a detail image. {And we aim to unroll this optimization algorithm as a trainable DNN to make the structure of our AUIF model to be more reasonable and efficient.}
	
	\subsection{Optimization Model}
	The base image is with low-frequency background information, so it is obtained by solving the following issue:
	\begin{equation}\label{equ:1}
	\begin{split}
	B^*&=\arg \min L_B \\
	&=\arg \min\frac{\theta_B}{2}\left\| {X - B} \right\|_F^2 + \mathop \sum \limits_{j = 1}^n \left\| {{g_j^B} * B} \right\|_F^2,
	\end{split}
	\end{equation}
	where $B^*$ is the decomposed base image, $X$ is the input image (including the infrared image $I$ and the visible image $V$), $g_j^B (j=1,\dots,n)$ are high-pass filters, $*$ represents the convolution operation, and $\theta_B$ is a tuning hyperparameter. In Eq.~(\ref{equ:1}), the first item is the data fidelity item and the second one is a regularization item to reduce the high-frequency of $B$. The detail image $D^*$ is with high-frequency texture and gradient information, and it can be similarly acquired by:
	\begin{equation}\label{equ:2}
	\begin{split}
	D^*&=\arg \min L_D \\
	&=\arg \min \frac{\theta_D}{2}\left\| {X - D} \right\|_F^2 + \mathop \sum \limits_{j = 1}^n \left\| {{g_j^D} * D} \right\|_F^2,
	\end{split}
	\end{equation}
	where $g_j^D (j=1,\dots,n)$ are low-pass filters, and $\theta_D$ is also a tuning hyperparameter.
	
	We use the gradient descent algorithm to solve the above issues. For model (\ref{equ:1}), the gradient of $L_B$ is calculated as:
	\begin{equation}\label{}
	\frac{{\partial L_B}}{{\partial B}} =  - \theta_B\left( {X - B} \right) + \mathop \sum \limits_{j = 1}^n \left(g_j^B\right)^\top  * \left( {{g_j^B} * B} \right).
	\end{equation}
	So the update rule of the base feature map is:
	\begin{equation}\label{equ:update_1}
	{B^{out}}\!=\!{B^{in}}\!-\!\eta_B\!\left[ { \mathop \sum \limits_{j = 1}^n \left(g_j^B\right)^\top\!*\!\left({{g_j^B}*{B^{in}}} \right)\!-\!\theta_B\!\left( {X\!-\!{B^{in}}} \right)}\!\right],
	\end{equation}
	where $\eta_B$ is the step size and $\left(g_j^B\right)^\top$ denotes the kernels of $g_j^B$ rotated by $180^\circ$. For the optimization of Eq.~(\ref{equ:2}), its update rule can be derived in the same way (omitted for space constraint).

	\subsection{Algorithm Unrolling}
	\subsubsection{BCL and DCL}
	Inspired by the work \cite{sreter2018learned}, we transform the optimization problem as a convolutional neural network. We replace the filters $\{g_j^B,g_j^D\}$ by convolutional units, and the update process Eq.~(\ref{equ:update_1}) can be rewritten as:
	\begin{equation}\label{equ:b_out}
	{B^{out}}\!=\!{B^{in}}\!-\!\eta_B \left[ Conv^B_2\left( {{Conv_1^B}\left( {B^{in}} \right)} \right)\!-\!\theta_B \left( {X\!-\!{B^{in}}} \right) \right],
	\end{equation}
	where $Conv_i^B (i=1,2)$ denotes the convolutional unit with a kernel of size $k$. In this paper, $k$ is set to 3. Notably, we also set the kernel of $Conv^B_2$ equals to that of $Conv^B_1$ rotated by $180^\circ$. For the proof of the convolution kernel setting in Eqs.~(\ref{equ:update_1})~and~(\ref{equ:b_out}), please refer to the supplementary material. Similarly, the update process of the detail feature map is implemented as:
	\begin{equation}\label{equ:d_out}
	{D^{out}}\!=\!{D^{in}}\!-\!\eta_D\!\left[ Conv^D_2 \left({{Conv^D_1}\left( D^{in} \right)} \right)\!-\!\theta_D \left( {X\!-\!D^{in}} \right) \right].
	\end{equation}
	
	In what follows, Eqs. (\ref{equ:b_out}) and (\ref{equ:d_out}) are named by Base Convolutional Layer~(BCL) and Detail Convolution Layer~(DCL), respectively. To keep the spatial size unchanged and prevent artifacts at the image edges \cite{DBLP:conf/iccv/ZhuPIE17}, the input of BCL and DCL is reflection-padded. To further enhance feature extraction capability, the output passes through a batch normalization layer and is activated by a parametric rectified linear unit (PReLU). It is worth pointing out that the filters $\{g^B,g^D\}$, step sizes $\{\eta_B,\eta_D\}$, and hyperparameters $\{\theta_B,\theta_D\}$ are pre-defined in traditional gradient descent algorithm. However, they are learnable in BCL and DCL.
	
	In conclusion, by transferring manually-designed filters to learnable deep convolution kernels, our model not only follows the traditional method but is more adaptable and reasonable with the parameters learnable.

	\subsubsection{Network Architecture}
	\begin{figure*}[!]
		\centering
		\includegraphics[width=\linewidth]{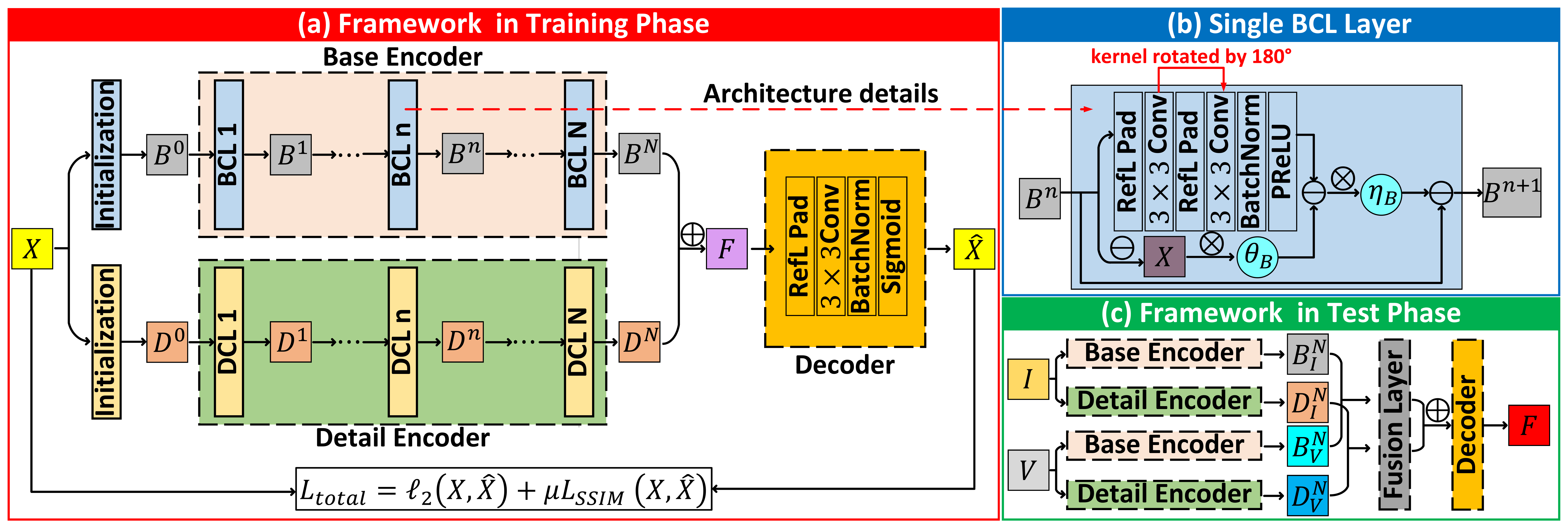}
		\caption{Illustration of the AUIF model. Note that the same color indicates the same network structure or the same type of feature map. (a): Network framework of AUIF in the training phase; (b): Display of a single BCL in the AUIF model, the same structure is also contained in DCL with only differences of parameters (omitted for space constraint); (c): Network framework of AUIF in the test phase. }
		\label{fig:Train_net}
	\end{figure*}
	Actually, both base and detail images can be regarded as feature maps of the source image $X$. We thus stack $N$ BCLs and DCLs as two encoders, simulating the iterative process of traditional optimization models, to extract base and detail feature maps. Then, an additional decoder is established whose inputs are the addition of two decomposed feature maps and output is the reconstructed source image $X$.
	
	The network architecture in training phase is displayed in Fig.~\ref{fig:Train_net}(a), a single BCL is shown in Fig.~\ref{fig:Train_net}(b) and a DCL has the same structure with different parameters. The number of channels $(c^{input},c^{output})$ for the first convolution units (i.e., $Conv_1^B$ and $Conv_1^D$) is $(1,C)$. As for the second convolutional units (i.e., $Conv_2^B$ and $Conv_2^D$), it is set as $(C,1)$. In this paper, $C$ is set to 64.
	There are no shared parameters in BCL and DCL. The input of base and detail encoders $B_0$ and $D_0$ are initialized by applying the blur and Laplacian filters to $X$, respectively. As for the decoder, it consists of a 3$\times$3 convolution unit, batch regularization layer, and the sigmoid function. Both the input and output channels of the convolution unit equal to 1. The role of the sigmoid function is to make the pixel values in the reconstructed image range from 0 to 1.
	
	\subsubsection{Loss Function}
	For the reconstruction loss that the AUIF net utilizes during the training phase, it is defined by
	\begin{equation}\label{equ:Loss_total}
	\begin{split}
	L_{total}  &=\ell_2(X,\hat{X}) + \mu L_{SSIM}(X,\hat{X})\\
	&=\left\| X-\hat{X} \right\|_2^2+ \frac{\mu}{2}\left(1 - SSIM(X,\hat X)\right),
	\end{split}
	\end{equation}
	where $\mu$ is the tuning parameter, $SSIM(\cdot,\cdot)$ is the structural similarity index \cite{wang2004image} which measures the similarity between two images. In Eq. (\ref{equ:Loss_total}), the $\ell_2$-loss ensures that the pixel intensity of the reconstructed image is close to the source image, while the $L_{SSIM}$ item makes the reconstructed image approximate the source image in terms of brightness, structure and contrast.
	
	\subsubsection{Test}
	In the test phase, we input the infrared and visible images in couples, then obtain the final fusion results.
	While after training, we have two well-trained encoders (decomposers) and a decoder (reconstructor).
	Here, $B_I^N$ and $D_I^N$ denote the base and detail feature maps of infrared images generated by the $N$th BCL and DCL (i.e., the output of base encoder and detail encoder), while $B_V^N$ and $D_V^N$ represent those of visible images.
	The specific workflow for the test phase is shown in Fig.~\ref{fig:Train_net}(c).
	
	Notably, different from the network in training phase, we need to set an extra fusion layer between the encoder and decoder to merge $\{B_I^N,B_V^N\}$ and $\{D_I^N,D_V^N\}$, respectively.
	$\Gamma(\cdot)$ is used to represent the pixel-wise feature map merging operator in the fusion layer, and it is defined by
	\begin{equation}\label{}
	\begin{split}
	B^N(x,y)=&\Gamma(B_I^N,B_V^N)=\alpha_I^B(x,y)\times B_I^N(x,y)\\
	+&\alpha_V^B(x,y)\times B_V^N(x,y),\\
	D^N(x,y)=&\Gamma(D_I^N,D_V^N)=\alpha_I^D(x,y)\times D_I^N(x,y)\\
	+&\alpha_V^D(x,y)\times D_V^N(x,y).
	\end{split}
	\end{equation}
	Three commonly used fusion strategies $\Gamma_i(\cdot) (i=1,2,3)$ are listed as follows: 1) Addition: $\displaystyle{\alpha_I^B=\alpha_V^B=\alpha_I^D=\alpha_V^D=1}$. 2) Average: The fusion weight can be manually designed, and the default setting is 0.5.
	3) $\ell_1$-attention Addition: Inspired by the work of \cite{li2018densefuse}, $\ell_1$-norm can reflect the salience degree of pixels. Thus we perform $\ell_1$-norm operation on the base and detail feature maps, based on which the merging weights can be calculated. The weights of base feature maps are defined by
	\begin{equation}\label{}
	\alpha _I^B = \frac{{\chi \left( {{{\left | {B_I^N} \right |}_1}} \right)}}{{\chi \left( {{{\left | {B_I^N} \right|}_1}} \right) + \chi \left( {{{\left| {B_V^N} \right|}_1}} \right)}},
	\alpha _V^B = 1-\alpha _I^B,
	\end{equation}
	where $\chi(\cdot)$ is the 3$\times$3 blur filter. The weights of detail feature maps $\alpha_I^D$ and $\alpha_V^D$ can be calculated in the same way.
	
	For the final selection of the fusion layer merging strategy, we will complete it on validation datasets. Please refer to section~\ref{sec:FusionLayer} for details.
	
	\section{Experiments} \label{sec:4}
	In this section, a series of experiments are conducted to study the behavior of our AUIF network. Experiments are implemented with Pytorch (version 1.3.1) on a computer with Windows 10 operating system, Intel Core i9-10900K CPU@3.70GHz, 64GB memory and NVIDIA GeForce RTX2080Ti GPU.

	\begin{table}[t]
		\centering
		\caption{Information of datasets in this paper.}
		\begin{tabular}{cccc}
			\toprule
			\multirow{2}{*}{Dataset} & Training   & \multicolumn{2}{c}{Validation} \\
			\cmidrule(r){2-2}\cmidrule(r){3-4}    & FLIR-Train & Urban-NIR     & Street-NIR     \\\midrule
			Illumination             & Day\&Night & Day           & Day            \\
			\# Image pairs           & 180        & 58            & 50             \\\midrule
			\multirow{2}{*}{Dataset} & \multicolumn{3}{c}{Test}                    \\
			\cmidrule(r){2-4}        & TNO        & FLIR-Test     & Country-NIR    \\\midrule
			Illumination             & Night      & Day\&Night    & Day            \\
			\# Image pairs           & 40         & 40            & 52             \\\bottomrule
		\end{tabular}
		\label{tab:dataset}
	\end{table}
	\subsection{Datasets and Metrics}
	\begin{figure*}[t]
		\begin{minipage}[b]{0.5\linewidth}
			\centering
			\includegraphics[width=\linewidth]{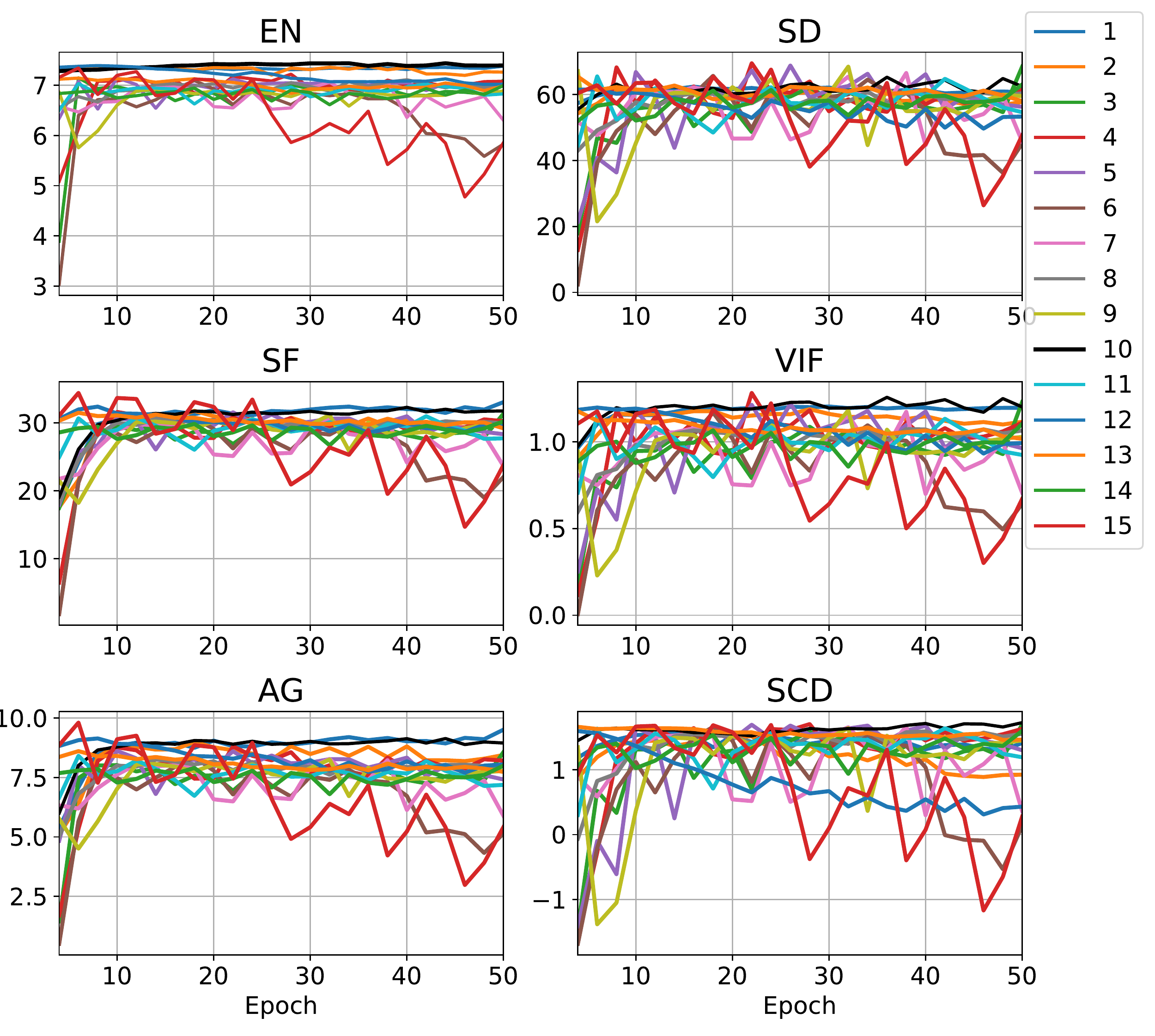}
			\par\vspace{0pt}
		\end{minipage}%
		\begin{minipage}[b]{0.5\linewidth}
			\centering
			\includegraphics[width=\linewidth]{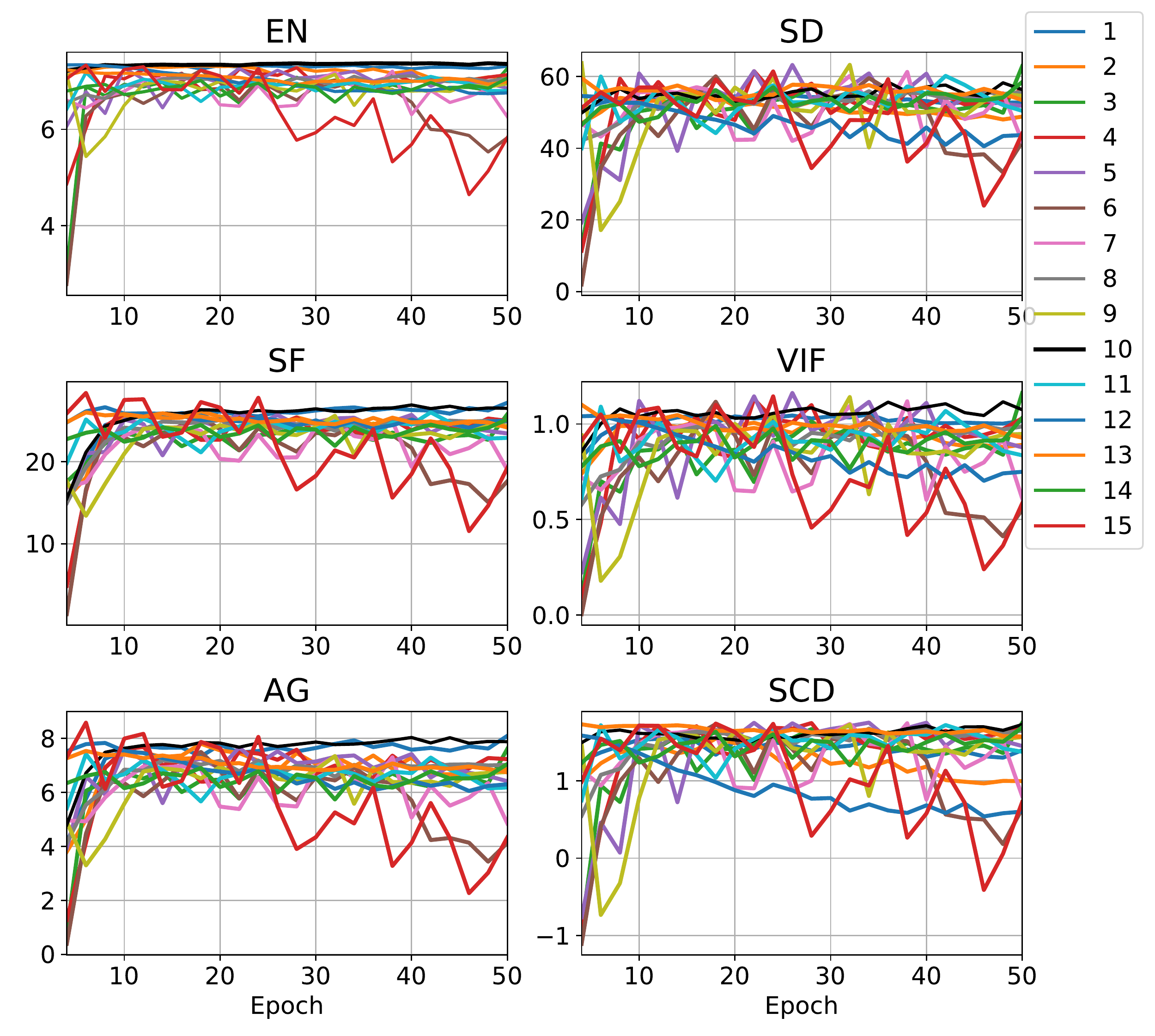}
			\par\vspace{0pt}
		\end{minipage}
		\caption{The results of verifying the number of BCL/DCL layers in validation datasets Urban-NIR (left) and Street-NIR (right). Lines in different colors represent the models containing different number of BCL/DCL layers.}
		\label{fig:val_result}
	\end{figure*}
	\subsubsection{Datasets}\label{sec:dataset}
	We use three IVIF datasets for our experiments: TNO \cite{TNO}, FLIR \cite{xu2020aaai}, and NIR \cite{brown2011multi}.
	The basic information is reported in Tab.~\ref{tab:dataset}. Note that the FLIR dataset is randomly divided into the training set and the test set.
	
	\subsubsection{Metrics}\label{sec:Metrics}
	Considering that there is no ground truth for the final fusion result, we select six image quality evaluation metrics: entropy (EN), standard deviation (SD), spatial frequency (SF), visual information fidelity (VIF),  average gradient (AG), and sum of the correlations of differences (SCD) to quantitatively describe the effectiveness of fusion.
	EN and SD measure the amount of information contained in fusion images. SF and AG reflect the detail and texture information of fusion images. VIF reports the consistency degree with the human visual system and SCD implies the agreement between source and fusion images. The higher the value is, the better quality of the fused results. More calculation details of the metrics can be found in \cite{ma2019infrared}.
	
	\begin{figure*}[!]
		\centering
		\subfigure{
			\includegraphics[width=\linewidth]{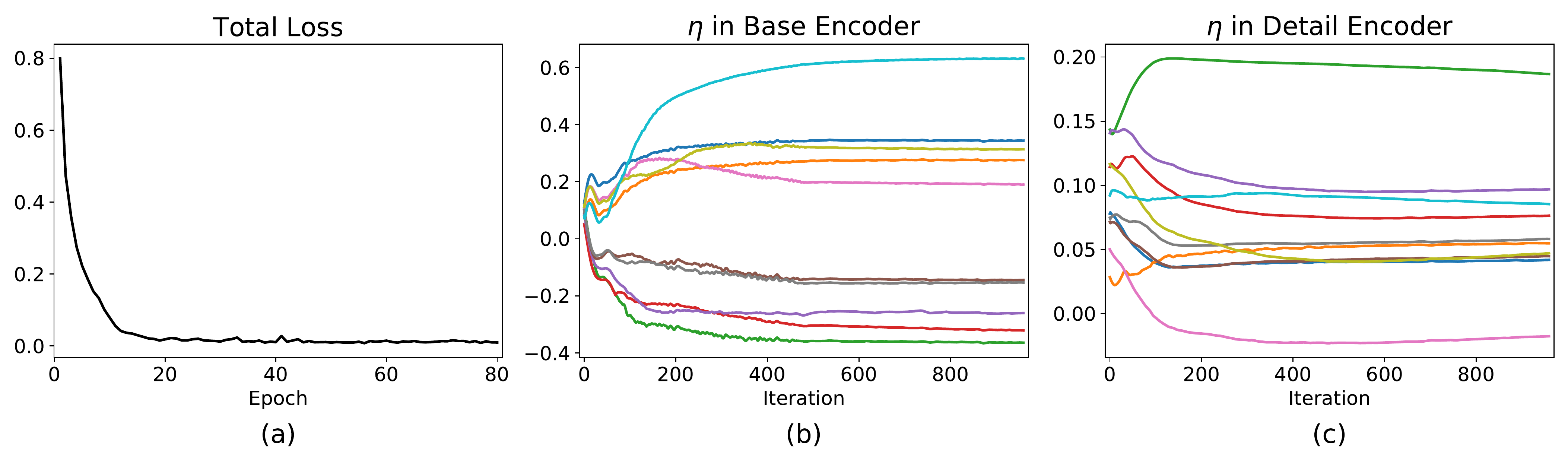}
		}
		\vspace{-2em}
		
		\subfigure{
			\includegraphics[width=0.8\linewidth]{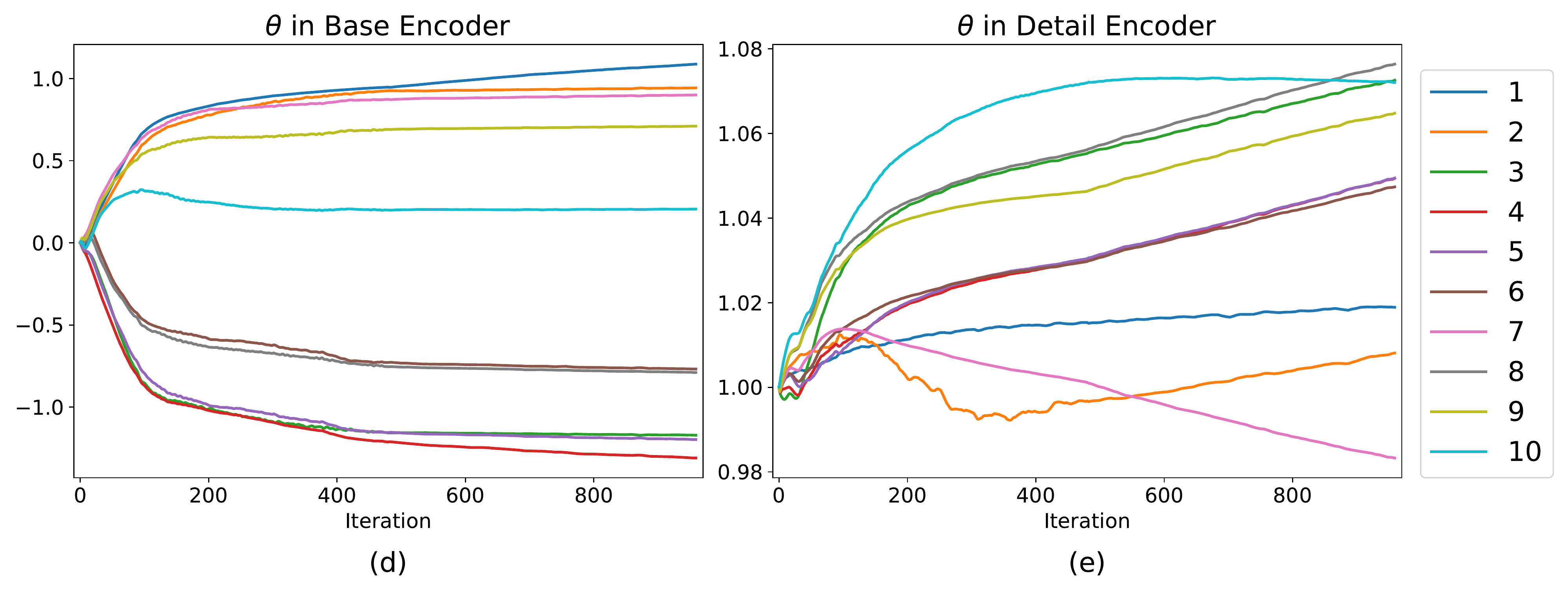}
		}
		\caption{Exhibition of training results. (a): The loss curve in 80 epochs. (b) \& (d): $\eta_B$ \& $\theta_B$ changes of each BCL layer in the Base Encoder. (c) \& (e): $\eta_D$ \& $\theta_D$ changes of each DCL layer in the Detail Encoder.
{ In (b)-(e), different colored lines represent different BCL or DCL layers (totally 10 layers in each subfigure).}}
		\label{fig:training_results} 
	\end{figure*}

	\subsection{Implementation Details and Network Configuration}
	In this experiment, we set the tuning parameter $\mu$ of Eq. (\ref{equ:Loss_total}) to 5. The AUIF network is trained over 80 epochs with a batch size of 32. The learning rate is $10^{-2}$ for the first 40 epochs and it is decreased to $10^{-3}$ for the rest epochs. The training samples are randomly cropped to 128$\times$128.
	
	For the learnable parameters $\eta$ and $\theta$ in Eq. (\ref{equ:b_out}) and (\ref{equ:d_out}), $\eta_b$ and $\eta_d$ are randomly initialized with a normal distribution $\mathcal N(0.1,0.03^2)$, while $\theta_b$ and $\theta_d$ are set to $10^{-3}$ and 1, respectively.
	The configuration of $\theta$ is related to $B^0$ and $D^0$. The initial detail feature map $D^0$ is generated by the Laplacian filter, and it is found that $D^0$ visually differs from the original image $I$. Thus, a larger $\theta_d$ is needed for the sake of data fidelity.
	In contrast, the initial base feature map $B^0$ is generated by the blur filter, and it is very similar to the source image. So, a smaller $\theta_b$ is needed to prevent from learning redundant features.
	At last, the number of layers $N$ is determined on validation sets. We vary $N$ from 1 to 15, and the quantitative results are reported in Fig. \ref{fig:val_result}. It is found that $N=10$ strikes the balance among six metrics on both Urban-NIR and Street-NIR datasets. The changing curves of training loss, $\eta$ and $\theta$ to the epoch number or iteration number are displayed in Fig.~\ref{fig:training_results}. It is shown that our network can converge rapidly with the above configuration.
	\subsection{Experimental Results}
	\subsubsection{Experiments on Fusion Layer}\label{sec:FusionLayer}
	Firstly, we need to choose a proper merging strategy for fusion layer on the validation sets.
	The results of the three strategies in the Street-NIR and Urban-NIR dataset are shown in Tab.~\ref{tab:VAl}.
	Obviously, the ``addition'' strategy reaches higher values on almost all metrics. Therefore, in the following experiments, we utilize the ``addition'' merging strategy in the fusion layer.
	\begin{table}[t]
		\centering
		\caption{Results on validation datasets. The best values are highlighted in bold.}	
		\begin{tabular}{cccc}
			\toprule
			          \multicolumn{4}{c}{\textbf{NIR Dataset. Scene: Street}}            \\
			Strategy &           Add           &          Ave           &  $\ell_1$-Att  \\ \midrule
			   EN    & \textbf{7.09$\pm$0.11}  &     6.82$\pm$0.06      & 6.89$\pm$0.05  \\
			   SD    & \textbf{53.62$\pm$2.02} &     36.07$\pm$0.95     & 36.92$\pm$2.22 \\
			   SF    & \textbf{24.78$\pm$1.24} &     17.06$\pm$0.47     & 17.45$\pm$1.31 \\
			  VIF    & \textbf{0.99$\pm$0.06}  &     0.68$\pm$0.05      & 0.54$\pm$0.05  \\
			   AG    & \textbf{7.09$\pm$0.56}  &     4.91$\pm$0.48      & 5.06$\pm$0.44  \\
			  SCD    & \textbf{1.57$\pm$0.14}  &     0.71$\pm$0.08      & 0.59$\pm$0.30  \\ \midrule
			           \multicolumn{4}{c}{\textbf{NIR Dataset. Scene: Urban}}            \\
			Strategy &           Add           &          Ave           &  $\ell_1$-Att  \\ \midrule
			   EN    &      7.04$\pm$0.24      & \textbf{7.11$\pm$0.05} & 7.07$\pm$0.11  \\
			   SD    & \textbf{59.09$\pm$2.49} &     41.86$\pm$1.78     & 41.52$\pm$1.63 \\
			   SF    & \textbf{28.99$\pm$1.33} &     20.47$\pm$0.25     & 20.73$\pm$0.91 \\
			  VIF    & \textbf{1.06$\pm$0.10}  &     0.78$\pm$0.06      & 0.82$\pm$0.07  \\
			   AG    & \textbf{8.11$\pm$0.58}  &     5.96$\pm$0.14      & 5.94$\pm$0.31  \\
			  SCD    & \textbf{1.44$\pm$0.16}  &     0.92$\pm$0.15      & 0.90$\pm$0.24  \\ \bottomrule
		\end{tabular}
		\label{tab:VAl}
	\end{table}
	\subsubsection{Ablation Experiments}\label{sec:AblationExperiment}
	\begin{table}[ht]
		\centering	
		\caption{Results of ablation experiments. The best values are highlighted in bold.}
		\begin{tabular}{ccccccc}
			\toprule
			                          \multicolumn{7}{c}{\textbf{NIR Dataset. Scene: Street}}                            \\
			           &      EN       &       SD       &       SF       &      VIF      &      AG       &      SCD      \\ \midrule
			  Exp. 1   &     6.91      &     50.98      &     23.23      &     0.92      &     6.74      &     1.38      \\
			  Exp. 2   &     7.02      &     50.68      &     24.63      &     0.92      &     6.79      &     1.26      \\
			  Exp. 3   &     6.87      &     41.75      &     23.65      &     0.76      &     6.10      &     0.73      \\
			  Exp. 4   &     6.85      &     41.64      &     23.62      &     0.75      &     6.08      &     0.72      \\
			  Exp. 5   &     6.52      &     40.16      &     22.57      &     0.66      &     5.39      &     0.54      \\
			  Exp. 6   &     6.33      &     36.76      &     21.95      &     0.59      &     4.93      &     0.34      \\
			  Exp. 7   &     6.58      &     44.25      &     21.87      &     0.78      &     5.77      &     0.89      \\
			  Exp. 8   &     6.85      &     48.64      &     21.51      &     0.81      &     6.05      &     1.19      \\
			   Ours    & \textbf{7.09} & \textbf{53.62} & \textbf{24.78} & \textbf{0.99} & \textbf{7.09} & \textbf{1.57} \\ \midrule
			                           \multicolumn{7}{c}{\textbf{NIR Dataset. Scene: Urban}}                            \\
			           &      EN       &       SD       &       SF       &      VIF      &      AG       &      SCD      \\ \midrule
			  Exp. 1   &     6.90      &     56.51      &     28.35      &     1.01      &     7.76      &     1.23      \\
			  Exp. 2   & \textbf{7.09} &     57.41      &     29.54      &     1.06      &     8.00      &     1.16      \\
			  Exp. 3   &     7.04      &     50.05      &     28.27      &     0.96      &     7.54      &     0.54      \\
			  Exp. 4   &     7.03      &     49.98      &     28.25      &     0.96      &     7.52      &     0.53      \\
			  Exp. 5   &     6.72      &     48.22      &     26.60      &     0.84      &     6.75      &     0.30      \\
			  Exp. 6   &     6.80      &     48.76      &     27.21      &     0.88      &     6.95      &     0.43      \\
			  Exp. 7   &     6.75      &     50.74      &     26.17      &     0.92      &     6.97      &     0.71      \\
			  Exp. 8   &     6.90      &     53.75      &     25.89      &     0.92      &     7.07      &     1.02      \\
			   Ours    &     7.04      & \textbf{59.09} & \textbf{29.99} & \textbf{1.06} & \textbf{8.11} & \textbf{1.44} \\ \bottomrule
		\end{tabular}	
		\label{tab:Ablation}
	\end{table}	
	\begin{figure}[ht]
		\centering	
		\includegraphics[width=0.5\linewidth]{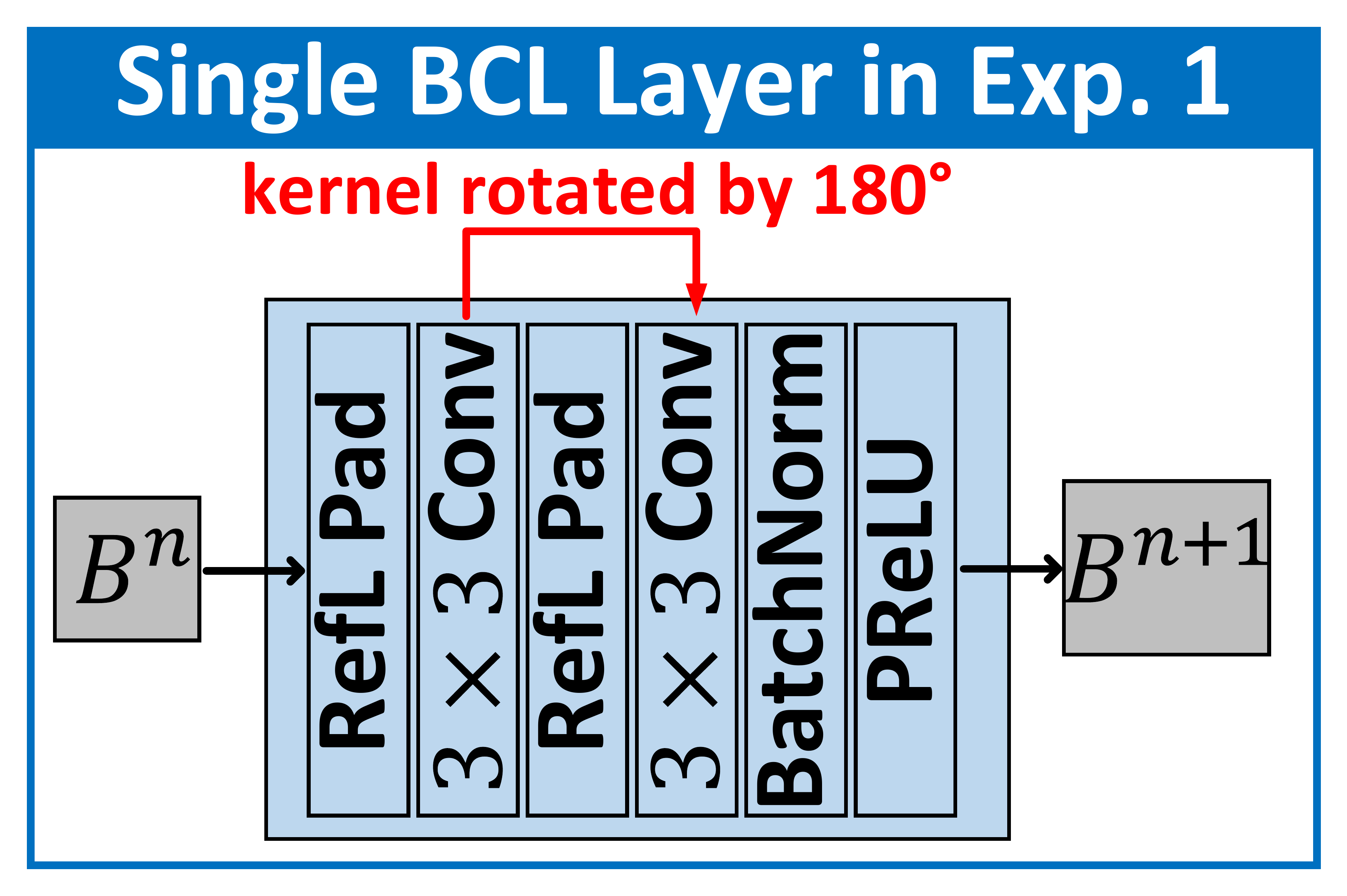}
		\caption{Diagram of single BCL in ablation Exp.~1. Note that DCL have the same structure, but with different parameters.}
		\label{fig:Ablation}
	\end{figure}	
	\begin{figure*}[!t]
		\centering	
		\includegraphics[width=\linewidth]{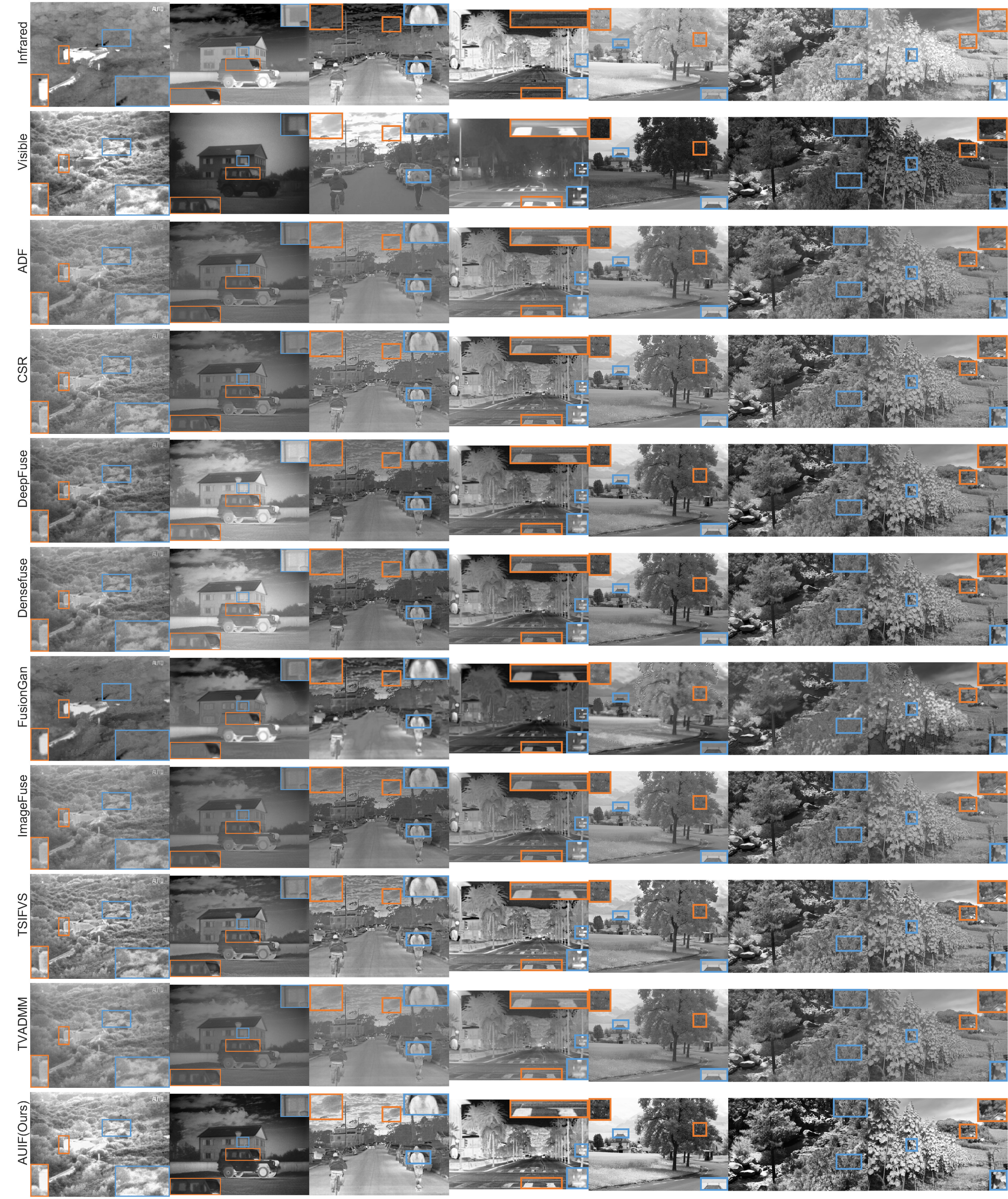}
		\caption{Exhibition of qualitative comparison results. From top to bottom: infrared images, visible images, results of SOTA methods and our method. {The blue and orange boxes represent the local areas that are respectively enlarged.}}
		\label{fig:Qualitative}
	\end{figure*}
	To verify the validity of some important modules in the AUIF model, we conduct ablation experiments as follows:
	
	\textbf{Effectiveness of algorithm unrolling:}	
	In \emph{Exp.~1}, we keep the number of layers in BCL/DCL unchanged, but the convolution units in each layer are merely connected end to end, without the calculations in Eqs. (3) and (4).
	That is, the overall frameworks of the model in training and testing are consistent with Fig.~\ref{fig:Train_net}(a) and Fig.~\ref{fig:Train_net}(c), but the structure of a single BCL/DCL
as shown in Fig.~\ref{fig:Train_net}(b) is replaced with that displayed in Fig.~\ref{fig:Ablation}.
	In \emph{Exp.~2}, initialization layers are deleted and training samples are directly input into the encoder in the training phase.
	
	\textbf{Effectiveness of feature decomposition:}	
	Then, we use traditional methods like the blur filter (\emph{Exp. 3}) or the optimization models (\emph{Exp. 4}) for two-scale decomposition, whose details are in the Appendix~\ref{sec:appendices}. Then the decoder is trained to accomplish the image reconstruction. In addition, we set the single base feature map $B^N$ (\emph{Exp. 5}) or single detail feature map $D^N$ (\emph{Exp. 6}) as the only input of the decoder to see whether a single decomposed feature map can achieve the fusion task.
	
	\textbf{Effectiveness of loss function:}	
	Finally, the loss function of the training network changes from $L_{total}=\ell_2+\mu L_{SSIM}$ to only $\ell_2$-loss (\emph{Exp. 7}) or only $\mu L_{SSIM}$ loss (\emph{Exp. 8}).
	
	We complete the ablation experiments on validation sets. Quantitative measurements for the fusion effectiveness in each ablation experiment and our model are demonstrated in Tab.~\ref{tab:Ablation}. Compared with the ablation experiment groups, our model performs best among cases in terms of almost all the metrics, which demonstrates the rationality of our model.
	
	\subsubsection{Qualitative Comparison}

	We compare our AUIF with eight SOTA methods, including
	ADF~\cite{bavirisetti2015fusion},
	CSR~\cite{liu2016image},
	DeepFuse~\cite{prabhakar2017deepfuse},
	Densefuse~\cite{li2018densefuse},
	FusionGAN~\cite{ma2019fusiongan},
	ImageFuse~\cite{li2018infrared},
	TSIFVS~\cite{bavirisetti2016two} and
	TVADMM~\cite{guo2017infrared}. Representative fusion results are shown in Fig.~\ref{fig:Qualitative}.
	
	We simply divide test samples into three categories: individuals, stuff and scenery.
	For individuals, our fusion images can reveal the specular lighting of targets, more details and clearer infrared radiation information. For stuff, our method can make interested ones be with sharpening edges. As a result, it is easy to distinguish the objects of interest from the background. For scenery, our results are clearer and have high contrast. Furthermore, the details of small objects are easier to observe.
	In conclusion, our model can retain both thermal radiation information and visible detail texture information.
	
	\subsubsection{Quantitative Comparison}
	\begin{table}[t]
		\centering
		\caption{Quantitative results of the SOTA methods in test datasets. The best and the second best values are highlighted by bold typeface and underline, respectively.}
		\begin{tabular}{lcccccc}
			\toprule
			                               \multicolumn{7}{c}{\textbf{Dataset: TNO image fusion dataset}}                                 \\
			Methods   &        EN        &        SD         &        SF         &       VIF        &        AG        &       SCD        \\ \midrule
			ADF       &       6.40       &       22.96       &       10.78       &       0.29       &       2.99       &       1.61       \\
			CSR       &       6.43       &       23.60       &       11.44       &       0.31       &       3.37       &       1.63       \\
			DeepFuse  & \underline{6.86} & \underline{32.25} &       11.13       & \underline{0.58} &       3.60       & \underline{1.80} \\
			DenseFuse &       6.84       &       31.82       &       11.09       &       0.57       &       3.60       &       1.80       \\
			FusionGan &       6.58       &       29.04       &       8.76        &       0.26       &       2.42       &       1.40       \\
			ImageFuse &       6.38       &       22.94       &       9.80        &       0.31       &       2.72       &       1.62       \\
			TSIFVS    &       6.67       &       28.04       & \underline{12.60} &       0.46       & \underline{3.98} &       1.68       \\
			TV-admm   &       6.40       &       23.01       &       9.03        &       0.28       &       2.52       &       1.60       \\
			Ours      &  \textbf{6.96}   &  \textbf{39.27}   &  \textbf{13.14}   &  \textbf{0.66}   &  \textbf{4.32}   &  \textbf{1.87}   \\ \midrule
			                               \multicolumn{7}{c}{\textbf{Dataset :FLIR image fusion dataset}}                                \\
			Methods   &        EN        &        SD         &        SF         &       VIF        &        AG        &       SCD        \\ \midrule
			ADF       &       6.80       &       28.37       &       14.48       &       0.34       &       3.56       &       1.39       \\
			CSR       &       6.91       &       30.53       &       17.13       &       0.37       &       4.80       &       1.42       \\
			DeepFuse  & \underline{7.21} & \underline{37.35} &       15.47       &       0.50       &       4.80       &       1.72       \\
			DenseFuse &       7.21       &       37.32       &       15.50       &       0.50       &       4.82       & \underline{1.72} \\
			FusionGan &       7.02       &       34.38       &       11.51       &       0.29       &       3.20       &       1.18       \\
			ImageFuse &       6.99       &       32.58       &       14.52       &       0.42       &       4.15       &       1.57       \\
			TSIFVS    &       7.15       &       35.89       & \underline{18.79} & \underline{0.50} & \underline{5.57} &       1.50       \\
			TV-admm   &       6.80       &       28.07       &       14.04       &       0.33       &       3.52       &       1.40       \\
			Ours      &  \textbf{7.45}   &  \textbf{47.58}   &  \textbf{19.94}   &  \textbf{0.64}   &  \textbf{5.94}   &  \textbf{1.88}   \\ \midrule
			                               \multicolumn{7}{c}{\textbf{Dataset: NIR image fusion dataset}}                                 \\
			Methods   &        EN        &        SD         &        SF         &       VIF        &        AG        &       SCD        \\ \midrule
			ADF       &       7.11       &       38.98       &       17.31       &       0.54       &       5.38       &       1.09       \\
			CSR       &       7.17       &       40.38       &       20.37       &       0.58       &       6.49       &       1.12       \\
			DeepFuse  &       7.30       &       45.82       &       18.63       &       0.68       &       6.18       &       1.37       \\
			DenseFuse & \underline{7.30} & \underline{45.85} &       18.72       &       0.68       &       6.23       & \underline{1.37} \\
			FusionGan &       7.06       &       34.91       &       14.31       &       0.42       &       4.56       &       0.51       \\
			ImageFuse &       7.22       &       42.31       &       18.36       &       0.61       &       5.92       &       1.22       \\
			TSIFVS    &       7.30       &       43.74       & \underline{20.65} & \underline{0.69} & \underline{6.82} &       1.19       \\
			TV-admm   &       7.13       &       40.47       &       16.69       &       0.53       &       5.32       &       1.09       \\
			Ours      &  \textbf{7.44}   &  \textbf{60.85}   &  \textbf{28.49}   &  \textbf{1.02}   &  \textbf{9.35}   &  \textbf{1.69}   \\ \bottomrule
		\end{tabular}	
		\label{tab:Quantitative}
	\end{table}
	\begin{figure*}[ht]
    	\centering	
    	\includegraphics[width=0.8\linewidth]{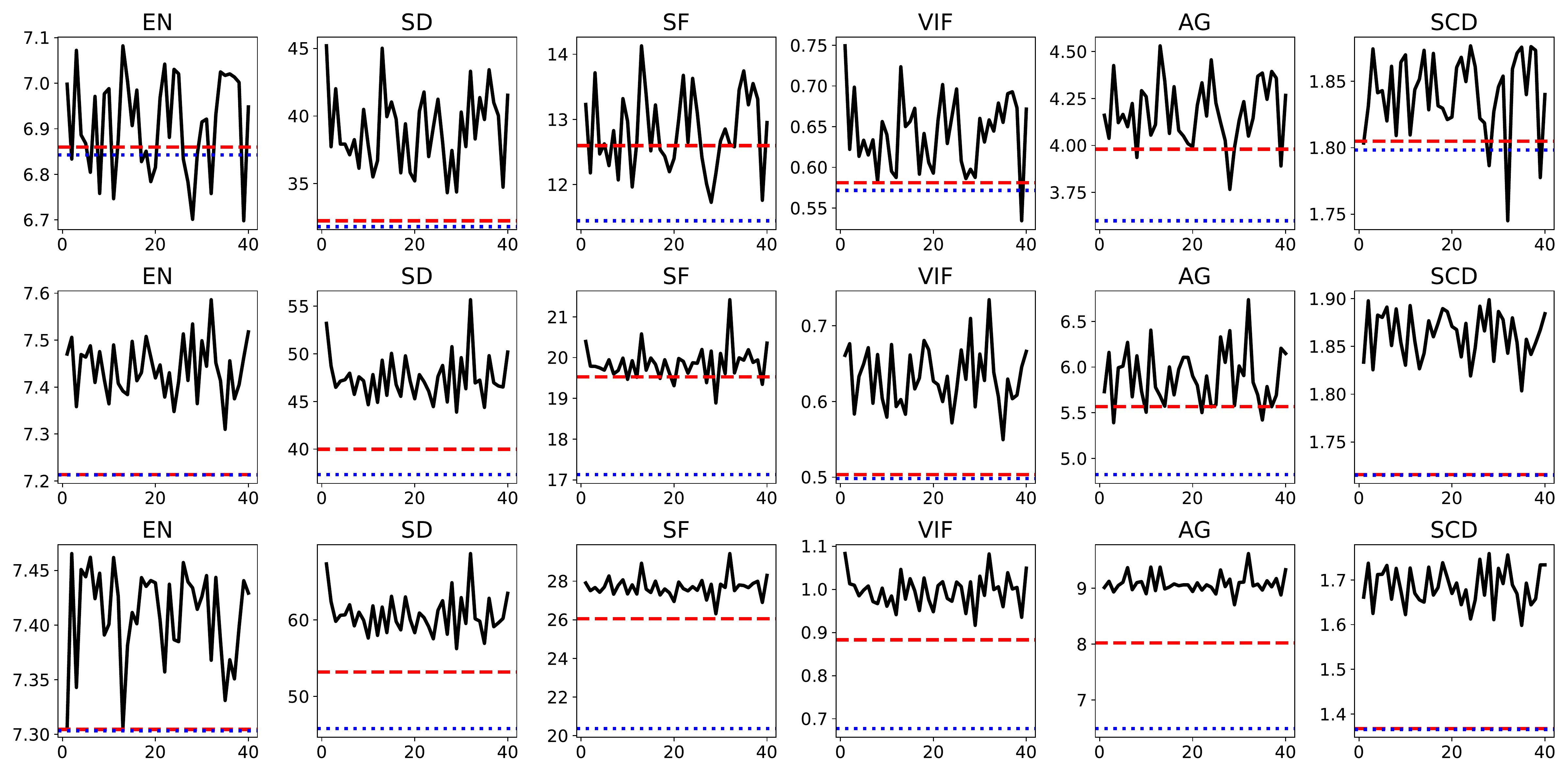}
    	\caption{The results of 40 parallel training. From top to bottom, the rows correspond to the results on TNO, FLIR, and NIR. The red dashed lines and the blue dotted lines represent the best and second best values among all competitors in each metric, respectively.}
    	\label{fig:Robustness}
	\end{figure*}
	Besides qualitative comparison, we use evaluation metrics to quantitatively measure the performance of all methods. The results of the three test datasets are exhibited in Tab.~\ref{tab:Quantitative}. It is shown that our method achieves excellent results on all test datasets with regard to all metrics. Conversely, others may perform well on a certain dataset with regard to part of metrics. It manifests that our method is suitable for the IVIF task under various illuminations and scenes.
	\subsubsection{Experiments on Robustness}
    To test the stability and reproducibility of our model, we repeatedly trained the AUIF network 40 times, and metric curves of the 40 models are shown in Fig. \ref{fig:Robustness}. To facilitate comparison, the top two values provided by eight competitors are set as baselines (see the red and blue dashed lines). It is observed that the AUIF network is an efficient fusion algorithm with high reproducibility.

    \subsubsection{Experiments on Computational Analysis}
    We then test the calculation time and the number of learnable parameters for AUIF and other SOTA methods. The results of the average calculation time in 40 repeated tests are shown in Tab.~\ref{tab:time} and the comparison for the number of learnable parameters in DL-based methods\footnote{ImageFuse~\cite{li2018infrared} is not added in the comparison of the number of learnable parameters, because it employs a pre-trained VGG-19 network.} is illustrated in Tab.~\ref{tab:Parameters}.
    For the average calculation time, our method can complete the fusion task in the second shortest time, which is only longer than that of the traditional method TSIFVS~\cite{bavirisetti2016two}. This proves the excellent calculation efficiency of AUIF.
    For learnable parameters in our method, it is significantly less than that in other DL-based methods, which demonstrates that the model-driven network structure based on algorithm unrolling can significantly reduce the number of learnable parameters and save computing resources.

    \begin{table}[t]
    	\centering
		\caption{The average calculation time in 40 repeated tests, and the unit of time is seconds.}
        \begin{tabular}{lrrr}
            \toprule
            \multirow{2}{*}{Methods} & \multicolumn{3}{c}{Average calculation time (s)} \\
             \cmidrule{2-4}
                & TNO (40 images)  & FLIR (40 images)   & NIR (52 images) \\
            \midrule
            ADF	&	12.57 	$\pm$ 	0.39 	&	6.61 	$\pm$ 	0.17 	&	9.54 	$\pm$ 	0.14 	\\
            CSR	&	2119.32 	$\pm$ 	9.80 	&	1320.06 	$\pm$ 	1.68 	&	1928.66 	$\pm$ 	2.45 	\\
            DeepFuse	&	3.85 	$\pm$ 	0.03 	&	3.60 	$\pm$ 	0.03 	&	5.41 	$\pm$ 	0.04 	\\
            DenseFuse	&	6.26 	$\pm$ 	0.04 	&	5.72 	$\pm$ 	0.03 	&	8.18 	$\pm$ 	0.03 	\\
            FusionGan	&	3.51 	$\pm$ 	1.50 	&	2.57 	$\pm$ 	0.02 	&	3.10 	$\pm$ 	0.04 	\\
            ImageFuse	&	149.83 	$\pm$ 	1.11 	&	96.04 	$\pm$ 	0.23 	&	134.97 	$\pm$ 	0.50 	\\
            TSIFVS	&	2.22 	$\pm$ 	0.09 	&	1.46 	$\pm$ 	0.01 	&	1.73 	$\pm$ 	0.01 	\\
            TV-admm	&	6.60 	$\pm$ 	0.05 	&	3.58 	$\pm$ 	0.02 	&	4.76 	$\pm$ 	0.02 	\\
            \midrule
            Ours	&	2.98 	$\pm$ 	0.03 	&	1.97 	$\pm$ 	0.02 	&	2.80 	$\pm$ 	0.02 	\\
            \bottomrule
        \end{tabular}
        \label{tab:time}
    \end{table}

    \begin{table}[t]
    \centering
	\caption{The learnable parameter numbers in DL-based methods, and the smallest value is highlighted in bold.}
        \begin{tabular}{cccc}
            \toprule
            FusionGan & DenseFuse & DeepFuse & AUIF (Ours)   \\
            \midrule
            925.63k   & 74.19k    & 89.16k   & \textbf{11.63k}\\
            \bottomrule
        \end{tabular}
    \label{tab:Parameters}
    \end{table}
    \begin{figure}[!t]
		\centering	
		\includegraphics[width=\linewidth]{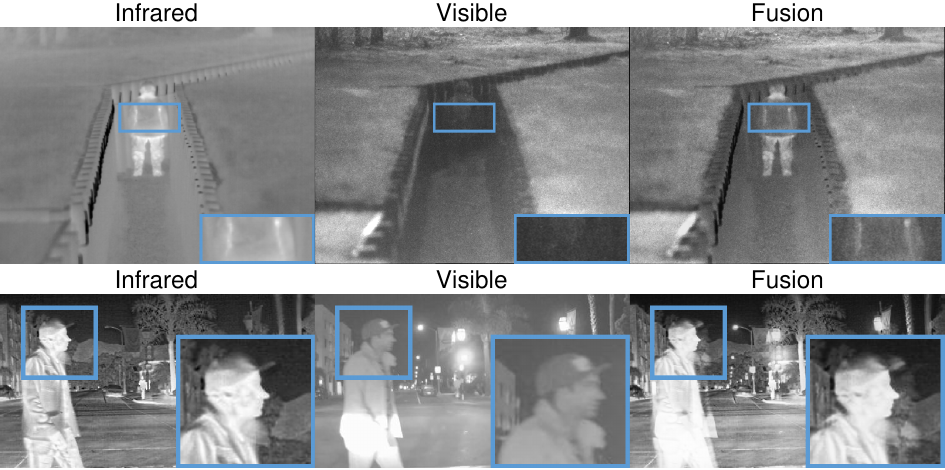}
		\caption{The representative samples for failure cases. {The blue boxes represent the enlarged local areas.}}
		\label{fig:failure}
	\end{figure}

    \subsubsection{Failure Cases and Weaknesses}
	
	Our method has a good performance on almost all datasets, while there are only two cases where larger errors may occur:
	a) Visible image is extremely dark, which may make objects hard to be distinguished.
	b) Unaligned images cause artifacts in the fused images.
	The representative samples are displayed in Fig.~\ref{fig:failure}.
	In addition, although AUIF can effectively handle IVIF tasks, it still has the following shortcomings:
	a) Images that are not well-aligned cannot be effectively fused.
	b) The initialization of some hyperparameters, i.e., $\eta$ and $\theta$, requires empirical design and cannot be obtained adaptively.
	c) AUIF is only designed for IVIF tasks, not for general image fusion tasks.
	The improvement will be completed in future work.

	\section{Conclusion}\label{sec:5}
	We design a novel deep fusion network by combining the {model-based prior information of traditional optimization models} and the strong feature extraction capability of deep neural networks.
	Firstly, two optimization models are established to complete the two-scale decomposition so that the decomposed feature maps could present different kinds of features. By setting different loss functions, detail feature maps with high-frequency information and base feature maps containing low-frequency information are disintegrated.
	Later, inspired by the idea of algorithm unrolling, the iteration steps of the optimization models can be extended to a neural network. The entire optimization model can be transformed into a neural network structure with a base encoder, a detail encoder, and a decoder. The encoders can decompose the image into base and detail feature maps, and the decoder is responsible for reconstructing the source image. Coefficients in the optimization model can be regarded as training parameters for our neural network.
	During the training phase, the initialized low-pass and high-pass filters are blur and Laplacian filters, respectively. Total loss function including $\ell_2$ and SSIM losses guarantee the performance of the network structure to reconstruct images.
	Subsequently, in the test phase, the two kinds of decomposed base (or detail) feature maps are input into the fusion layer to merge respectively and the final fusion image is the output of the decoder. The number of iterations (the number of neural network layers) and the fusion layer addition strategy are acquired from the validation datasets. In addition, the rationality of the module setting can be proved by various ablation experiments.
	Ultimately, numerous experiments conducted on TNO, FLIR, and NIR datasets demonstrate that our model can generate satisfactory fusion images with high reproducibility and computational efficiency.

	In the future, we hope to use the algorithm unrolling model for other multi-modal image fusion or image reconstruction tasks. In addition, we will also explore the combination of traditional and deep learning-based methods in computer vision to solve various tasks.	
	\begin{appendices}
		\section{Traditional Decomposition Methods.}\label{sec:appendices}
		The previously used decomposition methods mentioned in section~\ref{sec:AblationExperiment}~(Exp. 3 and 4) are the filter method and optimization model method. They are defined as follows:
		
		We denote $B^*$ and $D^*$ as the decomposed base and detail feature maps:
		\begin{itemize}
			\item The filter method is:
			\begin{equation}
			B^*=\psi(I),\ D^*=I-B^*,
			\end{equation}
			where $I$ is the source image and $\psi(\cdot)$ is a blur filter.
			\item The optimization method is:
			\begin{equation}
			\small
			\begin{split}
			B^*=\arg&\min \mathcal{F}\left( I,B\right) \\
			\mathcal{F}\left( I,B\right)=||I-B||_F^2&+\lambda(||g_x *B||_F^2+||g_y *B||_F^2)\\
			D^*&=I-B^*
			\end{split}
			\end{equation}
			where $*$ represents the convolution operation, $g_x=[-1,1]$ and $g_y=[-1,1]^T$.
		\end{itemize}
	\end{appendices}

	\ifCLASSOPTIONcaptionsoff
	\newpage
	\fi
	
	\bibliographystyle{IEEEtran}
	\bibliography{egbib}
	
	\begin{IEEEbiography}[{\includegraphics[width=1in,height=1.25in,clip,keepaspectratio]{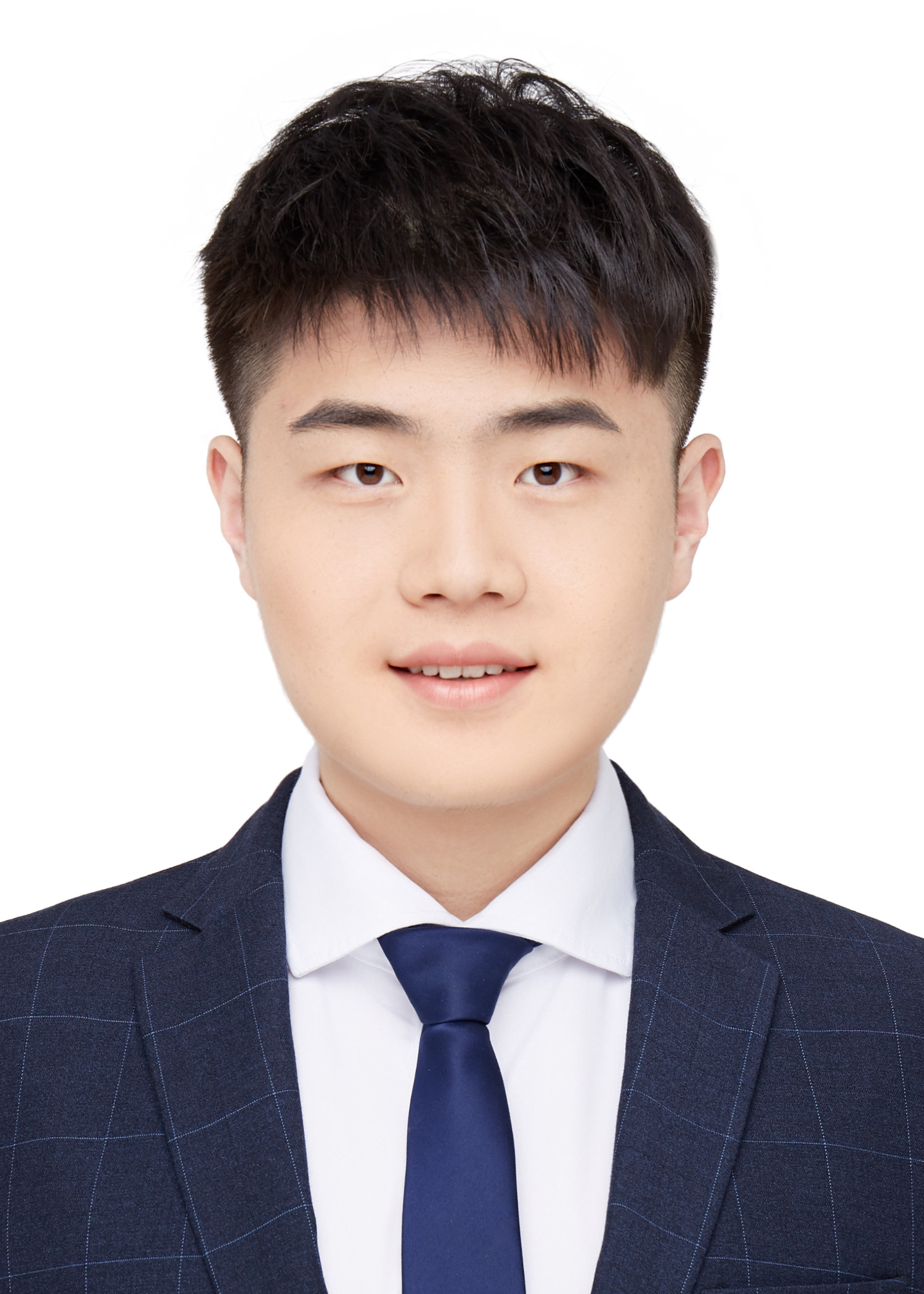}}]{Zixiang Zhao} is currently pursuing the Ph.D. degree in statistics with the School of Mathematics and Statistics, Xi’an Jiaotong University, Xi’an, China. His research interests include computer vision, deep learning and low-level vision.\end{IEEEbiography}
	
	\begin{IEEEbiography}[{\includegraphics[width=1in,height=1.25in,clip,keepaspectratio]{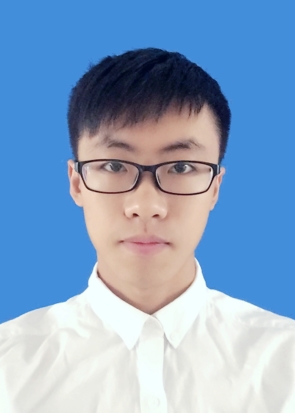}}]{Shuang Xu} is currently pursuing the Ph.D. degree in statistics with the School of Mathematics and Statistics, Xi'an Jiaotong University, Xi'an, China. His current research interests include Bayesian statistics, deep learning and complex network.
	\end{IEEEbiography}
	%
	\begin{IEEEbiography}[{\includegraphics[width=1in,height=1.25in,clip,keepaspectratio]{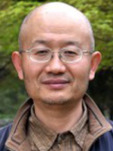}}]{Jiangshe Zhang} was born in 1962. He received the M.S. and Ph.D. degrees in applied mathematics from Xi'an Jiaotong University, Xi'an, China, in 1987 and 1993, respectively, where he is currently a Professor with the Department of Statistics. He has authored and co-authored one monograph and over 80 conference and journal publications on robust clustering, optimization, short-term load forecasting for the electric power system, and remote sensing image processing.
		
		His current research interests include Bayesian statistics, global optimization, ensemble learning, and deep learning.
	\end{IEEEbiography}
	
	\begin{IEEEbiography}[{\includegraphics[width=1in,height=1.25in,clip,keepaspectratio]{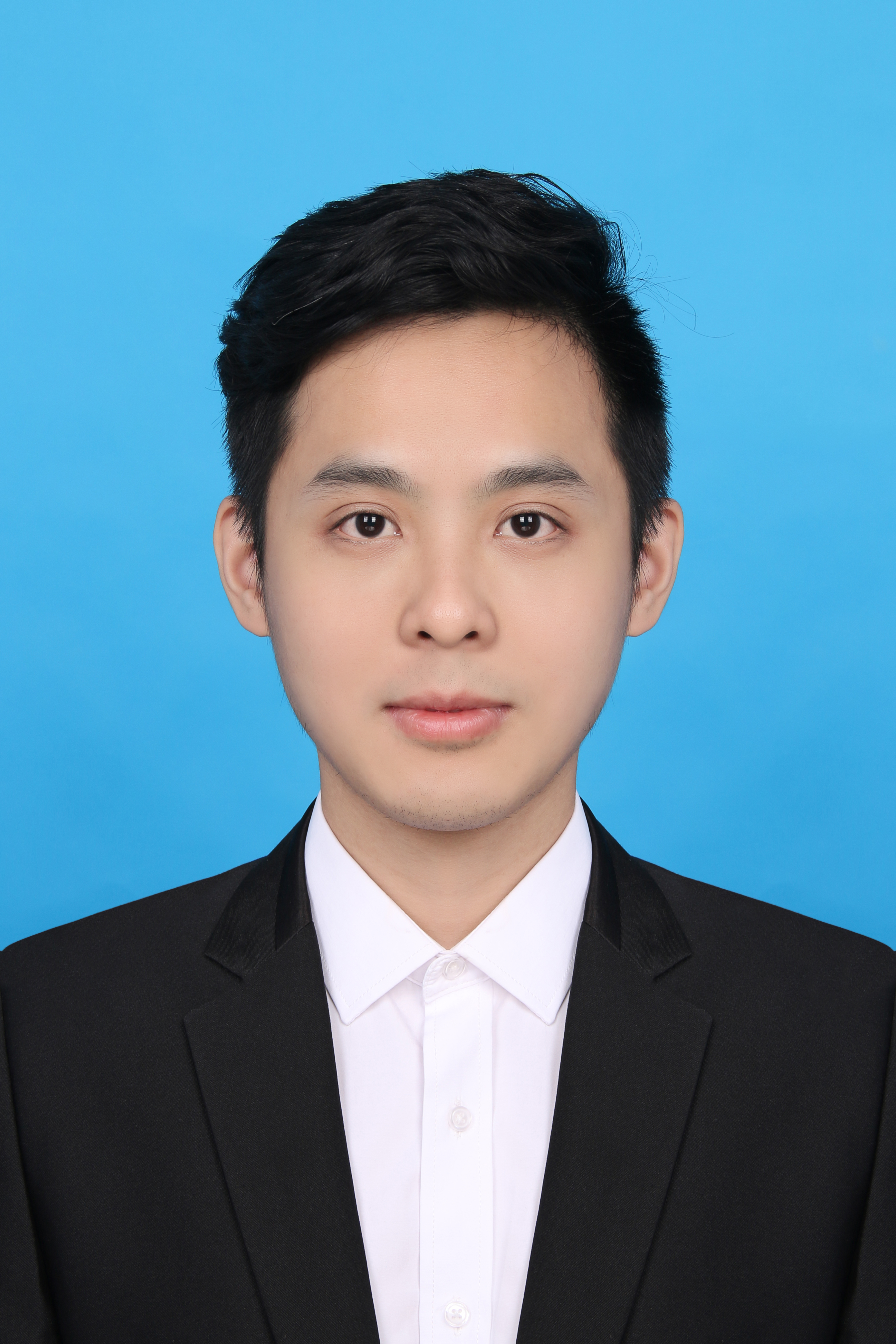}}]{Chengyang Liang} is currently pursuing the master's degree in statistics with the School of Mathematics and Statistics, Xi’an Jiaotong University, Xi’an, China. His research interests include transfer learning and machine learning.\end{IEEEbiography}
	%
	\begin{IEEEbiography}[{\includegraphics[width=1in,height=1.25in,clip,keepaspectratio]{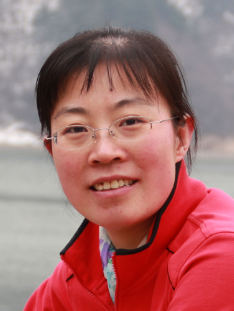}}]{Chunxia Zhang} received her Ph.D degree in Applied Mathematics from Xi'an Jiaotong University, Xi'an,  China, in 2010.
		
		Currently, she is a Professor in School of Mathematics and Statistics at Xi'an Jiaotong University. She has authored and coauthored about 30 journal papers on ensemble learning techniques, nonparametric regression, etc. Her main interests are in the area of ensemble learning, variable selection and deep learning.
	\end{IEEEbiography}
	
	\begin{IEEEbiography}[{\includegraphics[width=1in,height=1.25in,clip,keepaspectratio]{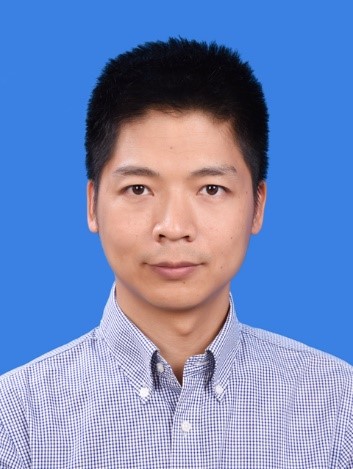}}]{Junmin Liu} (M'13)
		received the M.S. degree in computational mathematics from Ningxia University, Yinchuan, China, in 2009, and the Ph.D. degree in applied mathematics from Xi’an Jiaotong University, Xi’an, China, in 2013.
		
		He is currently an Associate Professor with the School of Mathematics and Statistics, Xi’an Jiaotong University. His current research interests include hyperspectral unmixing, remotely sensed image fusion, and deep learning.	
	\end{IEEEbiography}

\end{document}